\documentclass[journal]{IEEEtran}[10pt]
\IEEEoverridecommandlockouts
\usepackage{amsmath,graphicx}
\usepackage{amsfonts}
\usepackage{amsbsy}
\usepackage{amsmath,amssymb}
\usepackage{amsthm}
\usepackage{times}
\usepackage{graphicx}
\usepackage{enumerate}
\usepackage[usenames]{color}
\usepackage[dvips]{pstcol}
\usepackage{epstopdf}
\usepackage{cite}
\usepackage{epsfig}
\usepackage{psfrag}
\usepackage{xcolor}
\usepackage{bm}
\usepackage{epstopdf}
\usepackage{cite}
\usepackage{color}
\usepackage{cuted}
\usepackage{xcolor}
\usepackage{verbatim}
\usepackage{algorithm}
\usepackage{algorithmic}
\usepackage{booktabs}
\usepackage{colortbl}
\usepackage{caption}
\usepackage{subfig}

\usepackage{enumitem}
\usepackage[colorlinks =True,
            linkcolor  =blue,
            anchorcolor=green,
            citecolor  =blue]{hyperref} 
            
\def\K{\mathbb{K}}
\def\N{\mathbb{N}}
\def\T{\mathbb{T}}
\def\Z{\mathbf{Z}}

\def\g{\boldsymbol{g}}
\def\TH{\boldsymbol{\Theta}}
\def\th{\boldsymbol{\theta}}
\def\la{\boldsymbol{\lambda}}
\def\LA{\boldsymbol{\Lambda}}

\newtheorem{assumption}{Assumption}[section]

\newtheorem{theorem}{Theorem}[section]

\newtheorem{lemma}[theorem]{Lemma}
\def\BibTeX{{\rm B\kern-.05em{\sc i\kern-.025em b}\kern-.08em
		T\kern-.1667em\lower.7ex\hbox{E}\kern-.125emX}}
\begin{document}
	
	\title{Federated Reinforcement Learning for Resource Allocation in V2X Networks
	}
	\author{Kaidi Xu, Shenglong Zhou, and Geoffrey Ye Li, {\it IEEE Fellow}
    \thanks{Kaidi Xu and Geoffrey Ye Li are with the ITP Lab, Department of EEE, Imperial College London, UK. Shenglong Zhou is with the School of Mathematics and Statistics, Beijing Jiaotong University, China. Emails:  k.xu21@imperial.ac.uk, slzhou2021@163.com, geoffrey.li@imperial.ac.uk}
			\thanks{
				*Corresponding author: Shenglong Zhou. 
			}}

		\maketitle
		
		\begin{abstract}
			Resource allocation significantly impacts the performance of vehicle-to-everything (V2X) networks. Most existing algorithms for resource allocation are based on optimization or machine learning (e.g., reinforcement learning). In this paper, we explore resource allocation in a V2X network under the framework of federated reinforcement learning (FRL). 
			On one hand, the usage of RL overcomes many challenges from the model-based optimization schemes. 
			On the other hand, federated learning (FL) enables agents to deal with a number of practical issues, such as privacy, communication overhead, and exploration efficiency. 
			The framework of FRL is then implemented by the inexact alternative direction method of multipliers (ADMM), where subproblems are solved approximately using policy gradients and accelerated by an adaptive step size calculated from their second moments. 
			The developed algorithm, PASM, is proven to be convergent under mild conditions and has a nice numerical performance compared with some baseline methods for solving the resource allocation problem in a V2X network.
		\end{abstract}
		
		
		\section{Introduction}
		The V2X networks have attracted considerable research interest since they are capable of delivering many important services, e.g., road safety and traffic efficiency, and enable various applications in smart cities, autonomous driving, and intelligent transport systems \cite{shah20185g,peng2019vehicular, noor2022survey}. Entities, including vehicles and roadside units in V2X networks, communicate and cooperate with each other and thus result in the coexistence of vehicle-to-vehicle (V2V) and vehicle-to-infrastructure (V2I) communications on the same spectrums. Therefore, complex mutual interference and severe performance degradation may arise. To overcome such drawbacks, proper resource allocation schemes need to be developed. 
		
		It has been noted that resource allocation is usually formulated as an optimization problem, which however is NP-hard in general and lacks universal low complexity and effective solutions. There is an impressive body of work on developing traditional optimization model-based approaches for resource allocation in V2X networks \cite{zeng2019joint,ashraf2016dynamic,liang2017spectrum,liang2017resource,liang2018graph,mei2018latency}. For example, by considering the density and physical proximity of vehicles,  a decentralized algorithm has been proposed in \cite{ashraf2016dynamic} to optimize the transmission delay and successful transmission probability. In \cite{liang2017spectrum}, a joint optimal centralized spectrum sharing and power control method has been developed to maximize the V2I link sum rate while guaranteeing the reliability of the V2V links with delayed channel-state-information (CSI) feedback. Furthermore, based on the slowly-varying large-scale fading information, the sum ergodic capacity of V2I links with V2V link reliability has been optimized in \cite{liang2017resource}. Additionally, the graph partitioning tool has been adopted to categorize the highly interfering V2V links into different clusters to reduce computational complexity and signaling overhead in \cite{liang2018graph}. 
		However, due to the fast-varying channel conditions, it is usually hard to obtain global CSI, which limits the practical implementation of the traditional model-based resource allocation schemes in V2X networks. Traditional centralized solutions usually lack scalability in large-scale V2X networks.
		Machine learning has great potential to address these issues. 
		
		\subsection{Related works}
		Reinforcement learning (RL), as an effective tool in machine learning, has gained popularity in recent decades and has been extensively employed to provide distributed resource allocation solutions for V2X networks. For instance, in \cite{ye2019deep}, each vehicle is treated as an agent and makes decisions on sub-channel and transmitted power selection with limited transmission overhead. The distributed resource allocation scheme in \cite{liang2019spectrum} is based on the multi-agent RL (MARL) algorithm, which optimizes the V2I link sum rate and the V2V link payload delivery rate. The MARL algorithm is further enhanced in \cite{he2020resource} by graph neural networks. 
		In addition to the aforementioned value-based RL algorithms, some other policy-optimization-based RL algorithms, e.g., policy gradient (PG) \cite{sutton1999policy}, deterministic PG (DPG) \cite{silver2014deterministic}, are also employed to solve the resource allocation problems in V2X networks.
		For instance, in \cite{nguyen2019distributed}, deep DPG is employed to solve the power allocation in D2D-based V2V communications.
		In \cite{saikia2023proximal}, a proximal policy optimization based RL algorithm has been proposed to optimize the phase-shift matrix of the reconfigurable intelligent surface (RIS) in RIS-assisted full duplex 6G-V2X Communications.
		
		When it comes to the privacy issue, federated reinforcement learning (FRL), as a distributed learning scheme, integrating federated learning (FL) and RL, enables each agent to learn the knowledge beyond its observability without sharing raw data \cite{smarakoon2018federated,qi2021federated, lu2021dynamic, li2022federated}. In \cite{lu2021dynamic}, FRL trains agents for dynamic channel access and power control in a distributed manner while preserving user privacy and reducing communication overhead. Recently, a federated MARL scheme in \cite{li2022federated} optimizes the cellular sum rate and the reliability and delay requirements of V2V links, where the FL can address the limitation of partial observability and accelerate the training process. 
		
		It is known that many FL algorithms, e.g., FedAvg \cite{mcmahan2017communication} and FedProx \cite{li2020federated}, have been proposed based on the gradient descent scheme. A separate line of research develops FL algorithms using inexact ADMM \cite{fedpd,zhou2021communication,zhou2023federated,zhou2023fedgia}. 
		The FedGiA algorithm in \cite{zhou2023fedgia} integrates the gradient descent and inexact ADMM. It has been shown to have high communication efficiency, low computational complexity, and convergence under weaker conditions. 
		In addition, compared with value-based RL systems, we can use continuous optimization techniques to train the policy-optimization-based RL systems.
		The PG-based MARL algorithm is analyzed and connected with optimization problems in \cite{leonardos2022global}.
		Therefore, we adopt partial ideas from FedGiA to FRL and design a PG-based Admm with Second Moment (PASM) algorithm to improve the performance of FRL.
		
		\subsection{Contribution}
		We employ the framework of FRL to train the agents for sub-channel and transmit power level selection in a V2X network, where each V2V link is deemed as an agent and learns to optimize the V2I link sum rate and the V2V link packet delivery rate based on local observation in a distributed manner. 
		The FRL framework is then implemented by the inexact ADMM where subproblems are solved approximately using PG. 
		Our main contribution is threefold.
		\begin{itemize}[leftmargin=10pt]
			\item We formulate the spectrum-sharing resource allocation problem in V2X networks as a MARL system to train a distributed resource allocation scheme.
			Specifically, we consider two different metrics, i.e., the successful package delivery rate of V2V links and the weighted sum rate of all links, in the V2X networks, where the first metric focuses more on the long-term reward while the second metric focuses more on the instantaneous reward.
			\item In the training phase, we exploit the FL and PG and propose a PASM algorithm to train the proposed MARL system in an FL manner. 
			Specifically, the agent policy optimization problem can be formulated as an FL problem.
			Then, we exploit the inexact ADMM  to solve the FL problem, where the second moment is adopted to further improve the algorithmic performance. Such information has been widely used in some popular optimizers in deep learning, e.g., Adam \cite{kingma2014adam} and RMSProp\footnote{RMSprop is an unpublished adaptive learning rate algorithm proposed by Geoff Hinton in Lecture 6e of his Coursera Class.}. 			 
			Despite the challenge of establishing the convergence property for an algorithm to solve RL problems, we manage to show that the proposed method, PASM, can converge under mild conditions.
			\item We implement PASM in the considered V2X network and compare it with a FedAvg-based FRL algorithm and an independent PG algorithm. Simulation results show that  PASM can achieve better performance in terms of obtaining moving average rewards.
			
		\end{itemize}

		\subsection{Organization}
		The outline of this paper is organized as follows.
		Section \ref{system_model} introduces the system model of a considered V2X network. 
		Our proposed PASM algorithm is introduced in Section \ref{PASM_alg}.
		The corresponding resource allocation scheme based on PASM for the considered V2X network is then introduced in Section \ref{RA_alg}. In the last two sections, we present the simulation results and conclude the article.

		\section{System model and problem formulation} \label{system_model}

		\begin{figure}
			\centering
			\includegraphics[width=0.45\textwidth]{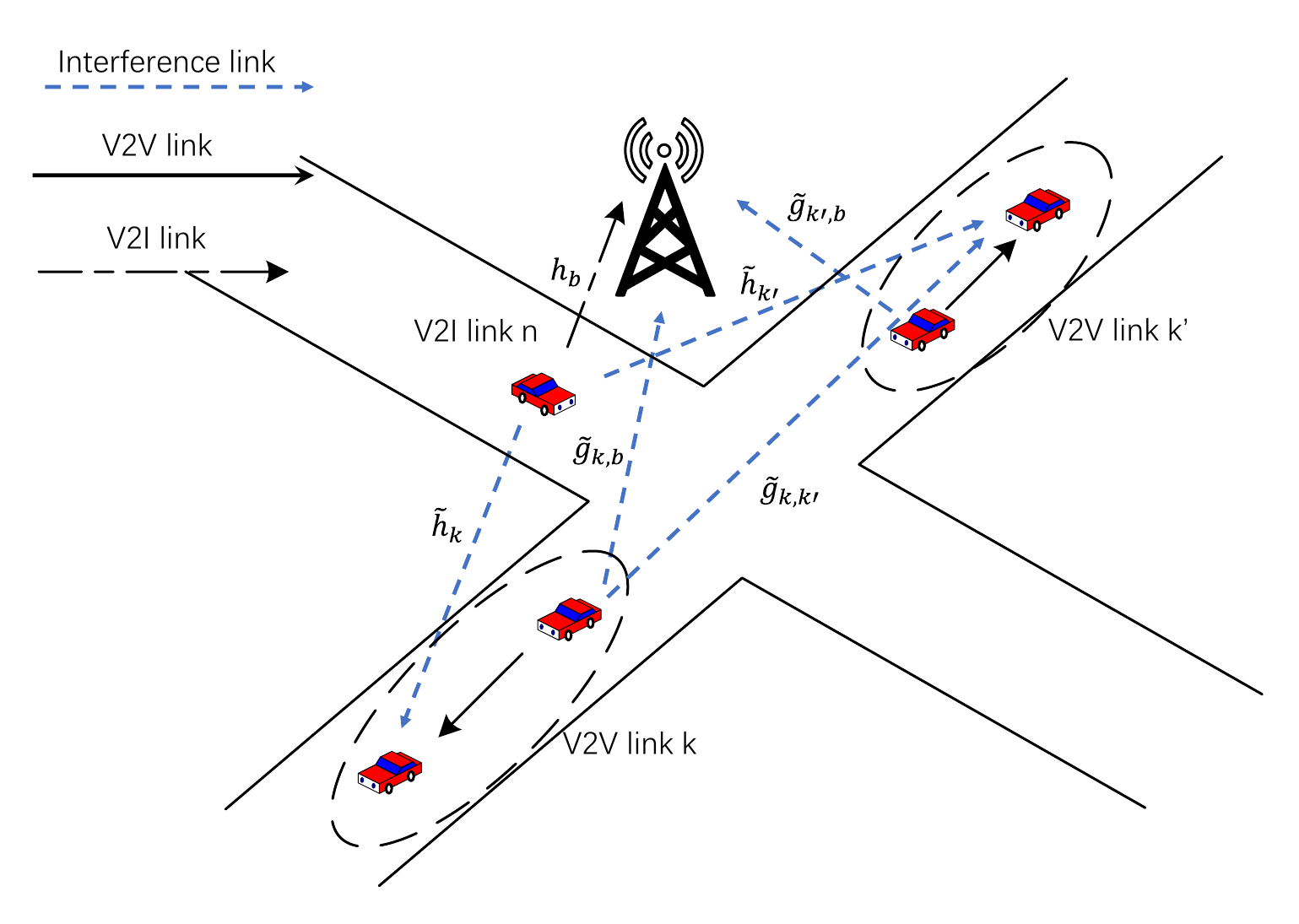}
			\caption{The V2X network diagram}
			\label{system}
		\end{figure}
		In this paper, we demonstrate the potential of FRL using resource allocation in V2X networks as an example.
		As shown in Fig. \ref{system}, we consider a single-antenna V2X network based on orthogonal frequency-division multiple access (OFDMA), where $N$ V2I links connect the vehicles and the base station (BS) and $K$ V2V links connect the neighboring vehicles.
		The V2I links support high-data-rate services and each of them is allocated with an orthogonal sub-channel. As a result,  the number of sub-channels matches the number of V2I links in the system under consideration. The V2V links are enabled by device-to-device (D2D) communication and reuse the uplink resource blocks allocated to V2I links to enhance system spectrum efficiency. We denote the set of V2I links as $\N=\{1,2,\ldots, N\}$, the set of V2V links as $\K=\{1,2,\ldots,K\}$, and the set of time slots as $\T=\{1,2,\ldots,T\}$. Assume that the $n$th sub-channel is allocated to the corresponding $n$th V2I link. The set of available sub-channels is denoted as $\N$.

		In time slot $t\in\T$, the signal-to-interference-plus-noise-ratio (SINR) of the $n$th V2I link can be expressed as,
		\begin{equation*}
			\gamma_{n,t}^i[n]=\frac{P_{n,t}^ih_{b,t}[n]}{\sum_{k\in{\K}}\delta_{k,t}[n]P_{k,t}^v[n]\tilde{g}_{k,b,t}[n]+\sigma^2},
		\end{equation*}
		where $h_{b,t}[n]$ denotes the channel power gain of the BS on the $n$th V2I link,
		$\tilde{g}_{k,b,t}[n]$ is the interference channel power gain from the transmitter of the $k$th V2V link to the $b$th BS on sub-channel $n$,
		$\sigma^2$ refers to the received Gaussian noise power,
		$P_{n, t}^i\leq P_{\max}^i$ and $P_{k,t}^v[n]\leq P_{\max}^v$ denote the transmit power of the $n$th V2I link and that of the $k$th V2V link on the $n$th sub-channel, respectively,
		and $\delta_{k,t}[n]$ is an binary indicator presenting the sub-channel allocation of V2V link $k$.
		If sub-channel $n$ is allocated to V2V link $k$, $\delta_{k,t}[n]=1$, otherwise $\delta_{k,t}[n]=0$.
		We also limit that each V2V link can occupy only one sub-channel, namely, $\sum_{n\in{\N}}\delta_{k,t}[n]\leq 1$ for all $k\in{\K}$ and $t\in{\T}.$
		The resulting achievable rate of V2I link $n$ in time slot $t$ is then given by,
		\begin{equation*}
			C_{n,t}^i=W\log(1+\gamma_{n,t}^i[n]),
		\end{equation*}
		where $W$ is the sub-channel bandwidth.
		
		For the $k$th V2V link, in time slot $t$, the corresponding SINR on sub-channel $n$ is given by,
		\begin{equation*}
			\gamma_{k,t}^v[n] = \frac{\delta_{k,t}[n]P_{k,t}^v[n]g_{k,t}[n]}{I_{k,t}^v[n] + P_{n,t}^i\tilde{h}_{k,t}[n]+\sigma^2},
		\end{equation*}
		where $I_{k,t}^v[n]=\sum_{k'\in{\K}, k'\neq k}\delta_{k',t}[n]P_{k',t}^v[n]\tilde{g}_{k',k,t}[n]$ denotes the interference power received by the receiver of V2V link $k$ in time slot $t$ from other V2V link transmitters on sub-channel $n$,
		$\tilde{g}_{k',k,t}[n]$ denotes the interference channel power gain from the transmitter of V2V link $k'$ to the receiver of V2V link $k$ on sub-channel $n$, and $\tilde{h}_{k,t}[n]$ denotes the interference channel power gain from the transmitter of V2I link $n$ to the receiver of V2V link $k$ on sub-channel $n$.
		Overall, we can express the corresponding achievable rate of V2V link $k$ in time slot $t$ as,
		\begin{equation*}
			C_{k,t}^v=\sum_{n\in{\N}}W\log(1+\gamma_{k,t}^v[n]).
		\end{equation*}
		The V2V links carry the safety-related information generated periodically, which needs to be delivered within a given time duration \cite{liang2019spectrum}.
		This V2V link transmission requirement is mathematically formulated as the delivery rate of packets of size $B$ within $T$ time slots,  
		\begin{equation*}
		p(\boldsymbol{\delta}_{k:},\boldsymbol{P}_{k:}^v):=	{\rm Pr}\Big(\Delta T\sum_{t\in{\T}}C_{k,t}^v\leq B\Big),~ \forall k \in {\K},
		\end{equation*}
		where {$\Delta T$} is the channel coherence time, {$\boldsymbol{\delta}_{k:}:=\{\delta_{k,t}[n]:t\in\T,n\in\N\}$}, and ${\boldsymbol{P}^v_{k:}:=\{P^v_{k,t}[n]:t\in\T,n\in\N\}}$.
		We consider two scenarios of the resource allocation for the V2X networks.
	\begin{itemize}[leftmargin=10pt]
\item 	Scenario I: one goal is to maximize the V2I link sum-rate and all V2V packet delivery rates $C_{k,t}^v$ by properly allocating the sub-channel and transmit power of V2V links with a given power control policy of V2I links, which is formulated as the following problem:
		\begin{equation}\label{p1}
			\begin{array}{cl}
				\max\limits_{\boldsymbol{\delta}_{k:},\boldsymbol{P}^v_{k:}, k\in\K} & \omega\sum\limits_{n\in{\N}}\sum\limits_{t\in {\T}} C_{n,t}^i + \sum\limits_{k\in\K} p(\boldsymbol{\delta}_{k:},\boldsymbol{P}_{k:}^v)\\[2ex]
				\text{s.t.}\quad &  \sum\limits_{n\in{\N}}\delta_{k,t}[n]\leq 1, \forall k,t\\[1.5ex]
				&\delta_{k,t}[n]\in \{0,1\}, \forall k,n,t \\[1ex]
				& 0\leq P_{k,t}^v[n]\leq P_{\max}^v, \forall k,n,t.
			\end{array}\tag{P1}
		\end{equation}

	\item	Scenario II: Another goal is to maximize the weighted sum rate of the V2V links and the V2I links in time slot $t\in\T$, which is a commonly used performance metric in many systems.
		The problem can be formulated as follows,
		\begin{equation}\label{p2}
			\begin{array}{cl} 
				\max\limits_{\boldsymbol{\delta}_{:t},\boldsymbol{P}^v_{:t}} & \omega\sum\limits_{n\in\N} C_{n,t}^i  +(1-\omega)\sum\limits_{k\in{\K}}C_{k,t}^v\\[2ex]
				\text{s.t.}\quad &  \sum_{n\in{\N}}\delta_{k,t}[n]\leq 1, \forall k \\[1.5ex]
				&\delta_{k,t}[n]\in \{0,1\}, \forall k,n\\[1ex]
				& 0\leq P_{k,t}^v[n]\leq P_{\max}^v, \forall k,n.
		\end{array}\tag{P2}
		\end{equation}
		
where for any $t\in\T$, ${\boldsymbol{\delta}_{:t}:=\{\delta_{k,t}[n]:k\in\K,n\in\N\}}$ and ${\boldsymbol{P}^v_{:t}:=\{P^v_{k,t}[n]:k\in\K,n\in\N\}}$.
\end{itemize}		

Note that we mainly focus on the resource allocation of the V2V links with a given V2I link power control policy.
		Therefore we fix the V2I link transmit power to its maximum level, i.e., $\{P_{n,t}^i=P_{\max}^i,\forall n,t\}$ in both considered scenarios.
		We aim to develop real-time distributed resource allocation schemes, which only require local observations for V2V links in these two scenarios.
		The V2V packet delivery rate in \eqref{p1} can be obtained after every $T$ time slot. On the other hand, the weighted achievable sum rate in \eqref{p2} is a short-term metric influenced by the global CSI and the resource allocation policy for each individual time slot $t$. Both problems are real-time sequential decision-making problems. We thus adopt the RL techniques to train distributed resource allocation schemes for problems \eqref{p1} and \eqref{p2}.

		\section{FRL via inexact ADMM and policy gradient}\label{PASM_alg}
		In this section, we will develop the algorithm based on the inexact ADMM.
		To begin with, we first introduce the considered cooperative MARL system for resource allocation in V2X networks.
		
		\subsection{Multi-agent policy gradient} \label{marl_env}
		A partially observable MARL system can be modeled as a partially observable Markov decision process (POMDP) with a tuple $\langle K, \mathbf{s}_t, \mathbf{a}_t^{(k)}, R_{t}^{(k)}, \mathbf{z}_{t}^{(k)}, P, O\rangle$, where $K$ is the number of agents, $\mathbf{s}_t$ is the environment state at time  $t$, $\mathbf{a}_t^{(k)}$ is the action at time  $t$ of agent $k$, 
		$\mathbf{z}_t^{(k)} = O(\mathbf{s}_t, k)$ is the local observation obtained by agent $k$, observation function $O(\cdot,\cdot)$ maps environment state $\mathbf{s}_t$ to a specific observation $\mathbf{z}_t^{(k)}$ of agent $k$, $R_t^{(k)}$ is the local reward received by agent $k$ from the environment, and $P(\mathbf{s}_{t+1}\mid\mathbf{s}_t, \mathbf{a}_t)$ is a transition probability from state $\mathbf{s}_{t}$ with action $\mathbf{a}_t$ to next state $\mathbf{s}_{t+1}$.

		The general idea of MARL is given as follows. At time step $t$, based on the local observation $\mathbf{z}_t^{(k)}$, agent $k$ selects an action $\mathbf{a}_t^{(k)}$ from the system's joint action $\mathbf{A}_t$ and receives a local reward $R_t^{(k)}$ from the environment. Then current state $\mathbf{s}_{t}$ transits to next state $\mathbf{s}_{t+1}$ with a transition probability $P(\mathbf{s}_{t+1}\mid\mathbf{s}_t, \mathbf{a}_t)$. 
		Subsequently, each agent $k$ obtains a new observation of the environment, $\mathbf{z}_{t+1}^{(k)}= O(\mathbf{s}_{t+1}, k)$. 
		In this paper, we investigate cooperative games, where all agents cooperate to improve the performance of the system.
		In other words, we consider a special case of MARL systems, i.e., the Markov Potential Game (MPG).
		
		Moreover, we take advantage of PG to cast our FRL framework.  It is noted that the PG-based method directly optimizes the policy of the agents to maximize the accumulative reward. More precisely, it maximizes the accumulative reward during a time period $T$ obtained by implementing the policy, $\pi_k(\mathbf{a}|\mathbf{z}_t^{(k)})$, which denotes the probability of performing action $\mathbf{a}$ when observing $\mathbf{z}_t^{(k)}$ for agent $k$. Denote $\boldsymbol{\Phi}$ the joint policy of all agents by\begin{equation}\label{decomposition}
			\boldsymbol{\Phi}(\mathbf{A}|\mathbf{s}_t) = \prod_{k=1}^K\pi_k(\mathbf{a}|\mathbf{z}_t^{(k)}).
		\end{equation}
		Then, given the static environment transition probability and the joint policy of all agents, in each episode a trajectory $\tau = \{\mathbf{s}_0,\mathbf{A}_0,$ $ \mathbf{s}_1, \mathbf{A}_1,\ldots, \mathbf{s}_T,\mathbf{A}_T\}$ of $T+1$ steps is sampled based on the policy and the environment.
		For agent $k$, the object is to maximize the expected accumulative reward with given policy $\boldsymbol{\Phi}$ over all possible trajectories, i.e., to maximize $$y_k(\boldsymbol{\Phi}) := \mathbb{E}_{\tau}(R^{(k)}(\tau)),$$
		where $R^{(k)}(\tau):=\sum_{t=0}^T R_t^{(k)}$ is the accumulative reward over trajectory $\tau$. According to \cite{leonardos2022global}, the PG for agent $k$ can be expressed as
		\begin{equation}\label{PG}
			\begin{aligned}
				\nabla_{\pi_k} y_k(\boldsymbol{\Phi}) 
				\approx \mathbb{E}_{\tau}\Big(R^{(k)}(\tau) \sum_{t=0}^T\nabla \log \pi_k(\mathbf{a}_t|\mathbf{z}_t^{(k)})\Big),
			\end{aligned}
		\end{equation}
		With the PG given in \eqref{PG}, the gradient ascent methods can be used to optimize the policy of all agents.
		
		Furthermore, by taking advantage of MPG, we can leverage the potential function of the MARL system, $\phi$, to formulate our FRL problem.
		According to Lemma 4.2 in \cite{leonardos2022global}, the stationary point of the potential function of the MARL system implies Nash policies of this MARL system.
		We thus aim to find the stationary point of the potential function of the MARL system, i.e., to find $\pi_k, \forall k$ that implies $\nabla_{\pi_k} \phi(\boldsymbol{\Phi})=\mathbf{0},\forall k$.
		On the other hand, gradient ascent methods can be used to find the stationary points without knowing the specific expression and derivatives of $\phi$ due to the equality of derivatives (cf. Proposition B.1 P2. in \cite{leonardos2022global}), given by,
		\begin{equation}\label{pg_eq}
			\nabla_{\pi_k} \phi(\boldsymbol{\Phi})=\nabla_{\pi_k} y_k(\boldsymbol{\Phi}),\forall k.
		\end{equation}
		For the sake of notation consistency, in the rest of this paper, we use the gradient of the potential function, $\nabla_{\pi_k} \phi(\boldsymbol{\Phi})$, to present the PG.
		\subsection{Inexact ADMM}

		Based on the discussion in Section \ref{marl_env}, we can formulate the PG-based FRL as an optimization problem to maximize the system potential function subject to the constraint that all agents share a common global policy model.
		The formulated FRL optimization problem is thus given by,
		\begin{equation}
			\begin{aligned}\label{FRL_PG_problem}
				\max_{\boldsymbol{\Theta},\boldsymbol{\theta}_c}\quad &\phi(\boldsymbol{\Phi})\\
				{\rm s.t.} \quad & \boldsymbol{\theta}_{k}=\boldsymbol{\theta}_{c},\forall k\in {\K},.
			\end{aligned}
		\end{equation}
		where $\boldsymbol{\theta}_c$ is the shared global model parameters and  $\boldsymbol{\Theta}:=(\boldsymbol{\theta}_1,\boldsymbol{\theta}_2,\ldots,\boldsymbol{\theta}_K)$ is a collection of all local trainable parameters $\{\boldsymbol{\theta}_k,k\in\K\}$. Here $\boldsymbol{\theta}_k$ denotes  agent $k$'s policy. In this context, we use a deep neural network to represent agent $k$'s policy. Therefore, $\pi_k$ is a function of $\boldsymbol{\theta}_k$, i.e., $\pi_k(\boldsymbol{\theta}_k)$ and thus the joint policy $\boldsymbol{\Phi}$ is a function of the collection of local parameters, i.e., $\boldsymbol{\Phi}:=\boldsymbol{\Phi}(\boldsymbol{\Theta})$. Therefore, hereafter, we denote
		$$\phi(\TH):=\phi(\boldsymbol{\Phi})=\phi(\boldsymbol{\Phi}(\boldsymbol{\Theta})).$$
		Note that \cite{agarwal2021theory, leonardos2022global} have proven the smoothness, i.e., the policy gradient Lipschitz continuity, of the expected value function in single-agent case and multi-agent case, respectively,
		which allows us to assume a gradient Lipschitz continuity on potential function $\phi$, namely,
		\begin{equation}\label{L_count}
			\|\nabla \phi(\boldsymbol{\Theta}_1)-\nabla \phi(\boldsymbol{\Theta}_2) \|\leq l\|\boldsymbol{\Theta}_{1}-\boldsymbol{\Theta}_{2}\|,
		\end{equation}
		where $\|\cdot\|$ is the Frobenius (or Euclidean) norm. We exploit the inexact ADMM to solve problem \eqref{FRL_PG_problem} in an FL manner. The augmented Lagrange function of  problem \eqref{FRL_PG_problem} is
		\begin{equation}\label{lagrange_function}
			\begin{aligned} 
				&L(\TH,\LA,\th_c) : = -\phi(\TH) + \sum_{k\in{\K}}L_k( \boldsymbol{\theta}_k,\boldsymbol{\lambda}_k,\boldsymbol{\theta}_c), \\
				& L_k( \boldsymbol{\theta}_k,\boldsymbol{\lambda}_k,\boldsymbol{\theta}_c):=  \boldsymbol{\lambda}_k^\top (\boldsymbol{\theta}_{k}-\boldsymbol{\theta}_{c}) + \frac{\rho}{2}\|\boldsymbol{\theta}_{k}-\boldsymbol{\theta}_{c}\|^2,
			\end{aligned}
		\end{equation}
		where $\rho>0$ and $\LA:=(\la_1,\la_2,\cdots,\la_k)$ is the collection of  all Lagrange multipliers. Then the inexact ADMM takes the framework as follows: given $(\TH^0,\LA^0)$, perform the following steps iteratively
		\begin{subequations}
\label{frame-ADMM}
\begin{align}
\TH^{j+1} & \approx {\rm argmin}_{\TH}~  L(\TH,\LA^j,\th_c^j), \label{theta_update}	\\[1ex]
\boldsymbol{\lambda}_k^{j+1}&= \boldsymbol{\lambda}_k^j + \rho(\boldsymbol{\theta}_{k}^{j+1} - \boldsymbol{\theta}_{c}^j),~~ k\in{\K},\label{lambda_update}\\[1ex]
\boldsymbol{\theta}_c^{j+1}&= {\rm argmin}_{ \boldsymbol{\theta}_c} \sum_{k\in{\K}}L_k( \boldsymbol{\theta}_k^{j+1},\boldsymbol{\lambda}_k^{j+1},\boldsymbol{\theta}_c),\label{shared_update}
\end{align}
\end{subequations}
for $j=0,1,2,\ldots$.
		To solve subproblem \eqref{theta_update} in the above scheme, we approximate $\phi(\TH)$ using first-order information, i.e., its PG $\nabla\phi(\TH)=(\nabla_{\th_1}\phi(\TH),\cdots,\nabla_{\th_K}\phi(\TH))$. Denote
\begin{equation}\label{def-theta-c-g}
			\begin{aligned}
			\TH_c&:=(\th_c,\th_c,\cdots,\th_c),\\[1ex]
			\g_k&:=-\nabla_{\th_k}\phi(\TH_c),~~\g_k^j:=-\nabla_{\th_k}\phi(\TH^j_c).
			 \end{aligned} \end{equation}		
			 From the above definition, $\g_k$ is the PG obtained by agent $k$ from the environment.  Now for each $k\in\K$, we can solve subproblem \eqref{theta_update} inexactly by  
		\begin{align}							
						\th_k^{j+1}&= \underset{\th_k}{\rm argmin} ~\th_k^\top\g_k^j + \frac{r_k}{2}\|\th_k-\th^j_c\|^2+ L_k(\th_k,\la_k^j,\th_c^j)\nonumber\\
\label{original_pg}			 &=\boldsymbol{\theta}_{c}^j - \frac{1}{\rho+r_k}(\boldsymbol{\lambda}_k^j + {\g}_k^j),
			\end{align}	
		where $ r_k \in(0,l]$ is a non-negative constant.
		The last step in \eqref{frame-ADMM} is the aggregation step  calculated by 
		\begin{equation}\label{agg_step}
			\boldsymbol{\theta}_c^{j+1} = \frac{1}{K}\sum_{k\in{\K}} \mathbf{u}_k^{j+1},
		\end{equation}
		where $\mathbf{u}_k^{j+1}$ is the temporary variables to be aggregated from agent $k$, which is updated locally at agent $k$ before the aggregation step, 
		\begin{equation}\label{z_update}
			\mathbf{u}_k^{j+1} = \boldsymbol{\theta}_k^{j+1}+\frac{1}{\rho}\boldsymbol{\lambda}_k^{j+1}.
		\end{equation}
		Unlike FL, FRL requires agents to interact with the environment to obtain the training data and estimate the PG. The PG, $\nabla\phi(\TH_c)$, is obtained from the interactions between the agents and the environment.

		\subsection{Second Moment}
		To further improve the learning process, the second moment is adopted to generate an adaptive step size,
		which has been proven effective in widely used optimizers, e.g., Adam \cite{kingma2014adam} and RMSprop. 
		In the above inexact ADMM algorithm, the Lagrange multipliers in \eqref{lambda_update} in fact contain the accumulated gradient information. 
		To some extent, the update of $\mathbf{u}_k^{j+1}$ in  \eqref{z_update} plays the role like gradient descent. One can treat $1/\rho$ as the stepsize and $\la_k^{j+1}$ as the direction. 
		This allows us to integrate the second moment into  \eqref{z_update}. 
		To proceed with that, for an initialized $\beta>0$ and $\mathbf{v}_{c}^0={\bf0}$, the server estimates the second moment by 
		\begin{equation}\label{moment_est}
			\mathbf{v}_{c}^{j+1} = \beta\mathbf{v}_{c}^{j}+\frac{1}{K}\sum_{k\in\K}(1-\beta){\boldsymbol{\lambda}}_k^{j+1}\odot {\boldsymbol{\lambda}}_k^{j+1},
		\end{equation}
		where $\beta$ is a moving average constant and $\odot$ is the Hardamard product. Then update \eqref{z_update} is modified as 
		\begin{equation}\label{new_pg}
			\mathbf{u}_{k}^{j+1}=\boldsymbol{\theta}_{k}^{j+1} + \frac{1}{\rho\Big(\sqrt{\mathbf{v}_{c}^{j+1} }+\epsilon\Big)} \odot \boldsymbol{\lambda}_k^{j+1},
		\end{equation}
		where $\epsilon$ is a small value to prevent zero denominators. 
		Here $1/\sqrt{\mathbf{v}}\in\mathbb{R}^N$ is a vector with the $n$th entry being $1/\sqrt{v[n]}$.

		\begin{algorithm}
			\caption{PASM: PG-based inexact ADMM using the second moment for RFL} 
			\begin{algorithmic}[1] \label{FRL-PG-ADMM}
				\STATE	 \textbf{Initialize}: ${\th_{c}^0= \mathbf{0},\mathbf{v}_{c}^0 = \mathbf{0}}$,  and {$ \boldsymbol{\lambda}_k=\mathbf{0}, \forall k\in{\K}$}, and proper hyper-parameters ${J>0}$, ${\rho>0}$, ${\epsilon\in(0,1)}$, ${\beta\in(0,1)}$, and ${r_k\in(0,l], \forall k\in{\K}}$.	\\
				
				\FOR{episode index $j = 0, 1, \ldots, J$} 
				\STATE \texttt{--Local gradient estimation--} 
				\FOR{each agent $k \in {\K}$}
				\STATE Updates its local model by $\boldsymbol{\theta}_{k}^{j} = \boldsymbol{\theta}_{c}^{j}$.
				\STATE Samples a trajectory $\tau_{k}^j$ based on current policy $\pi_k(.;\boldsymbol{\theta}_{k}^j)$ and calculate the PG of the current episode based on \eqref{PG} to derive $\g_k^{j} =  -\nabla_{\boldsymbol{\theta}_k} \phi(\TH_c^j)$.
				\STATE Update  $(\th_k^{j+1},\la_k^{j+1})$ by \eqref{original_pg} and \eqref{lambda_update} and send them to the sever.	
				\ENDFOR 
				\STATE \texttt{--Global aggregation--} 
				\STATE The server updates the global parameter $\th_c^{j+1}$ by \eqref{moment_est}, \eqref{new_pg} and \eqref{agg_step}, and broadcasts it to all agents.
				\ENDFOR 		
			\end{algorithmic}
		\end{algorithm}
		
		The resulting framework is summarized in Algorithm \ref{FRL-PG-ADMM}.
		All agents update their models and aggregate in each episode.
		In Step 3, the local update of the agents starts to be performed. 
		In Step 6, each agent interacts with the environment, obtains the experience trajectory of the current episode, and calculates the PG.
		Then in Step 7, each agent updates the local model parameters and the Lagrange multipliers.
		Step 10 aggregates the knowledge of all agents at the server.
		
	\subsection{Convergence analysis}	
		
	 Before analyzing the convergence of the PASM algorithm, we need the following assumptions. 
		\begin{assumption}\label{ass}
			Suppose that 1) $\phi$ is gradient Lipschitz continuous, i.e. (\ref{L_count}), 2) $\phi$ is bounded from below, i.e., {$\phi>-\infty$}, and 3) $\max_{k\in\K}\|\nabla_{\th_k}\phi(\TH_c)\|_{\infty}\leq 1-\epsilon$.
		\end{assumption}	 
		The first two assumptions are commonly used to established the convergence in optimization. The third assumption can be guaranteed if we set up $|R_t^{(k)}|$ to have a small upper bound. Indeed,    if we choose tiny reward $|R_t^{(k)}|$, then $R^{(k)}(\tau)=\sum_{t=0}^T R_t^{(k)}$ can be sufficiently small, resulting in small $\|\nabla_{\pi_k} y_k(\boldsymbol{\Phi})\|_{\infty}$ by \eqref{PG} and so is $\|\nabla_{\th_k}\phi(\TH_c)\|_{\infty}$ by \eqref{pg_eq}. Here $\|{\bf x}\|_{\infty}=\max_{n\in\N}|x[n]|$ is the infinity norm. 
        To analyze the convergence, we need the following lemma proved in Appendix A.
			\begin{lemma}\label{convergence-lemma}
			Suppose $\max_{k\in\K}\|\nabla_{\th_k}\phi(\TH_c)\|_{\infty}\leq 1-\epsilon$, then $\|\mathbf{v}_{c}^{j}\|_{\infty}<(1-\epsilon)^2$ for any $j=0,1,2\cdots$.
		\end{lemma}
		Based on the above lemma, we establish the following convergence guarantee of the PASM algorithm, where $\Z^{j+1}:=(\TH^{j+1},\LA^{j+1},\th_c^{j})$.
		\begin{theorem}\label{convergence}
			Suppose Assumptions \ref{ass} hold and choose  ${\rho\geq 10l, \epsilon\in (0.5,1)}$. Then 1) sequence ${\{L(\Z^{j})\}}$ is non-increasing and converges, 2) $\lim_{j\to \infty} \|\Z^{j+1}-\Z^{j}\|=0$, and 3) the policy gradient is vanishing eventually, i.e., $\lim_{j\to \infty} \|\sum_{k\in\K}{\g}_k^j\| = 0$. 	
		\end{theorem}
		The above theorem is proved in Appendix B.
  The conditions given in Theorem \ref{convergence} are sufficient but unnecessary, which indicates that there is no need to set up parameters strictly satisfying these conditions for the algorithm to converge in the numerical experiments.
		

		\section{PASM for resource allocation in V2X networks} \label{RA_alg}

\begin{figure}[th]
			\centering
			\includegraphics[width=0.45\textwidth]{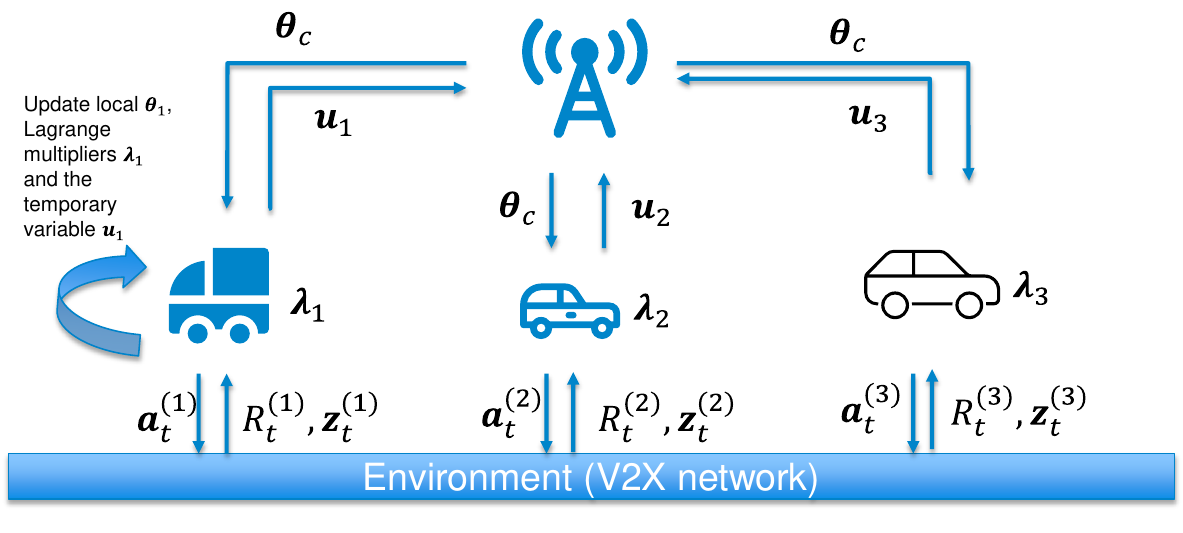}
			\caption{PASM workflow in a V2X network with 3 vehicles.}
			\label{workflow}
		\end{figure}

		In this section, we apply PASM to the resource allocation problems in the considered V2X network. Fig. \ref{workflow} depicts an example of PASM in a V2X network with 3 agents. In the considered FRL-for-V2X-network setting, the base station, which provides V2I link services to the vehicles, is regarded as the central training server in the FRL framework, while the local agent models are trained and deployed at each vehicle. Each vehicle updates the local model parameters and local Lagrange multipliers based on its collected local experience at the end of each episode. During the aggregation phase, each vehicle uploads its local model parameters and Lagrange multipliers to the base station via V2I links. After collecting these messages from the vehicles, the base station aggregates all this local model information to obtain a global model and then broadcasts it to all the vehicles via V2I downlinks.
		
		To apply the proposed FRL algorithm, we first formulate the resource allocation problem as a Multi-Agent Reinforcement Learning (MARL) system. Specifically, each V2V link is treated as an agent in the RL framework. Each agent maintains a policy deep neural network to make decisions. In both considered scenarios described in Section \ref{system_model}, the agents have the same observable information. Thus, we use the same observation space but different reward functions for the two scenarios. Since each V2V link determines its sub-channel selection and transmits power level, the action space of each agent, denoted as $k$, is defined as ${(\ell_k, P_k^v)|\ell_k \in \mathbb{N}, P_k^v \in \mathcal{P}}$, where $\ell_k$ denotes the selected sub-channel index and $\mathcal{P}$ denotes the available discrete transmit power levels defined as $\mathcal{P} = {23, 10, 5, -100}$dBm in the sequel.\footnote{The discrete action space here can be extended to a continuous action space easily as in the continuous PG algorithm \cite{silver2014deterministic}.}

		For practical implementation, each agent only has local observations and we do not use interference CSI as a part of the local observation.
		Specifically, the local observation of agent $k$ includes the V2I link channel power gains over all sub-channels, its V2V link channel power gains over all sub-channels, its received interference power over all sub-channels in the last time slot,  relative position $\mathbf{q}_k$ between the transmitter and the receiver of V2V link $k$,  velocities  $vel_k^t,vel_k^r$ of the transmitter and the receiver of V2V link $k$, the remaining time budget $T_k$,  remaining payload $B_k$ to be transmitted and the agent index $k$.
		Formally, we have  
		\begin{equation*}
			\begin{aligned} 
				\mathbf{z}_k^t=\Big\{h_{b,t}[n],g_{k,t}[n],I_{k,t-1}^v[n]+P_{n,t-1}^i\tilde{h}_{k,t-1}[n],\forall n\in{\N}, &\\
				\mathbf{q}_k,vel_k^t,vel_k^r, T_k^t, B_k^t, k\Big\}.&
			\end{aligned}
		\end{equation*}
		Note that local observation $\mathbf{z}_k^t$ is easy to obtain at each vehicle. We thus assume that there is no observation collection delay for the agents.
		Therefore, the agents are able to make real-time decisions for the V2V links.
		
		For Scenario I, we aim to maximize the successful delivery rate of the V2V links and the sum rate of the V2I links.
		We thus use a common reward for all agents defined as 
		\begin{equation}R_t = \omega\sum_{n\in{\N}}C_{n,t}^i+\sum_{k\in{\K}}D_{k,t}+\sum_{k\in{\K}}U_k. \label{reward}\end{equation} In \eqref{reward}, $D_{k,t}$ is a stimulus used to encourage the agent to transmit packets if the remaining packet data size is positive,
		\begin{equation*}
			D_{k,t} = \begin{cases}
				C_{k,t}^v& \text{if } B_k>0,\\
				0, & \text{otherwise}.
			\end{cases}
		\end{equation*}
		$U_k$ is determined after the episode ends and it indicates whether V2V link $k$ successfully delivers all its packets in the episode,
		\begin{equation*}
			U_k=\begin{cases}
				\Omega, & \text{if } B_k\leq 0\text{ at the end of the episode},\\
				0, & \text{otherwise}. 
			\end{cases}
		\end{equation*}
		Here, $\Omega$ is a large positive constant to encourage agents to successfully deliver all their data packets.
		
		For Scenario II, we aim to maximize the weighted achievable sum rate of all V2I links and V2V links.
		Therefore, we directly use the weighted achievable sum rate as the reward as follows,
		\begin{equation}
			R_t = \omega\sum_{n\in{\N}}C_{n,t}^i+(1-\omega)\sum_{k\in{\K}}C_{k,t}^v. \label{reward2}
		\end{equation}

		\section{Simulation results}\label{simulation_result}
		\begin{table}
			\centering
			\caption{Simulation Parameters}
			\begin{tabular}{||m{0.22\textwidth} | m{0.18\textwidth}||} 		
				\hline
				Parameter & Value \\ [0.5ex] 
				\hline\hline
				Carrier frequency & 2GHz \\ 
				\hline
				Bandwidth & 4MHz \\
				\hline
				BS antenna height & 25m  \\
				\hline
				BS antenna gain & 8dBi \\
				\hline
				BS receiver noise figure & 5dB  \\  
				\hline
				Vehicle antenna height & 5m \\
				\hline
				Vehicle antenna gain & 3dBi\\
				\hline
				Vehicle receiver noise figure & 9dB\\
				\hline
				Vehicle speed & 10-15m/s \\
				\hline
				Vehicle drop and mobility model & Urban case of A.1.2 in \cite{3GPP_TR_36.885_V14.0.0}\\
				\hline
				V2I transmit power $\{P_{n,t}^i\}$ & 23dBm\\
				\hline
				V2V transmit power $\{P_{k,t}^v\}$ & [23,10,5,-100]dBm\\
				\hline
				Noise power $\sigma^2$& -114dBm\\
				\hline
				V2V package delivery time & 100ms\\
				\hline
				V2V link packet size & 1060 bytes\\
				\hline		
				Channel fast-fading updating time & 1ms\\
				\hline
			\end{tabular}	\label{sys_para}
		\end{table}
		In this section, we demonstrate the performance of PASM for resource allocation in a V2X network through computer simulation. Our simulation environment follows the urban case in
		Annex A of \cite{3GPP_TR_36.885_V14.0.0}.
		We consider $N$ V2I links and $K$ V2V links in the V2X network, where the V2V links are formed by each vehicle and its neighbors. 
		We test the performance of the proposed algorithm under different $(N,K)$ pairs.
		The simulation parameters are summarized in Table \ref{sys_para}.
		
		The policy deep neural network for each V2V link consists of three fully connected hidden layers with $500$, $250$, and $120$ neurons, respectively.
		The rectified linear unit (ReLU) function is used as the activation function in the input and three hidden layers. The output layer is connected to a softmax function so that the final output is a probability distribution of the action. Each training episode consists of $100$ time slots. For Scenario I, we set the V2I link sum rate weight $\omega=0.01$ and the V2V successful delivery reward $\Omega=0.5$, as the V2V package delivery rate is more important. We set the hyper-parameters of the PASM algorithm as $\rho=1000$, $\epsilon=10^{-2}$, $\alpha=1$, and $\beta=0.999$. 
		For Scenario II, we set $\omega=0.1$, $\rho=500$, and other parameters the same as those in Scenario I.
		
		We compare our PASM algorithm with the independent PG algorithm and the FedAvg-based FRLPG algorithm \cite{mcmahan2017communication}. 
		Both algorithms employ Adam \cite{kingma2014adam} optimizer to update the local policy deep neural networks and share the same neural network structure with the PASM algorithm. The learning rate of the PG algorithm and the FRLPG algorithm is set as $10^{-4}$ and $10^{-3}$, respectively\footnote{
			We set a slower learning rate for the PG algorithm because a slightly larger one (e.g., $10^{-3}$) makes the PG algorithm fail to learn a good policy. In addition, we find that the ADAM optimizer and RMSprop optimizer have very similar performance in the FRLPG and the Independent PG algorithms. Therefore, we only show the results of ADAM optimizer in the simulation.}.
		We also use two additional baselines. The random resource allocation scheme randomly chooses the sub-channel and transmits the power level, which is a lower bound of the system performance. The centralized maxV2V in \cite{liang2019spectrum} provides an upper bound of Scenario I by an exhaustive search scheme.\footnote{We only apply this baseline to the case of $(N,K)=(4,4)$, as the brute-force method has an extremely high complexity when the number of agents increases.}
		
		\begin{figure}[htbp]
			\centering
			 \includegraphics[width=.47\textwidth]{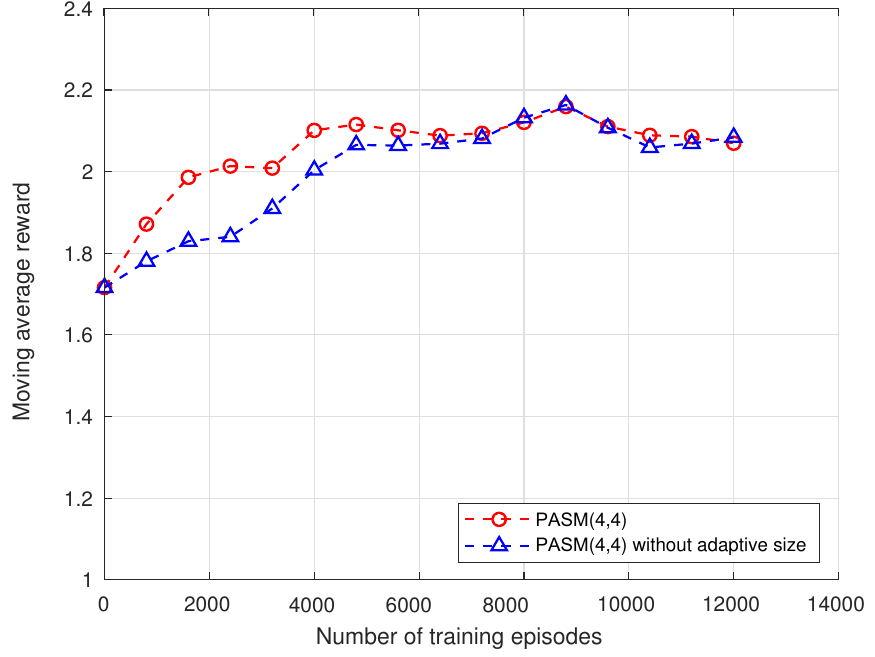}\\[2ex]
			 \includegraphics[width=.47\textwidth]{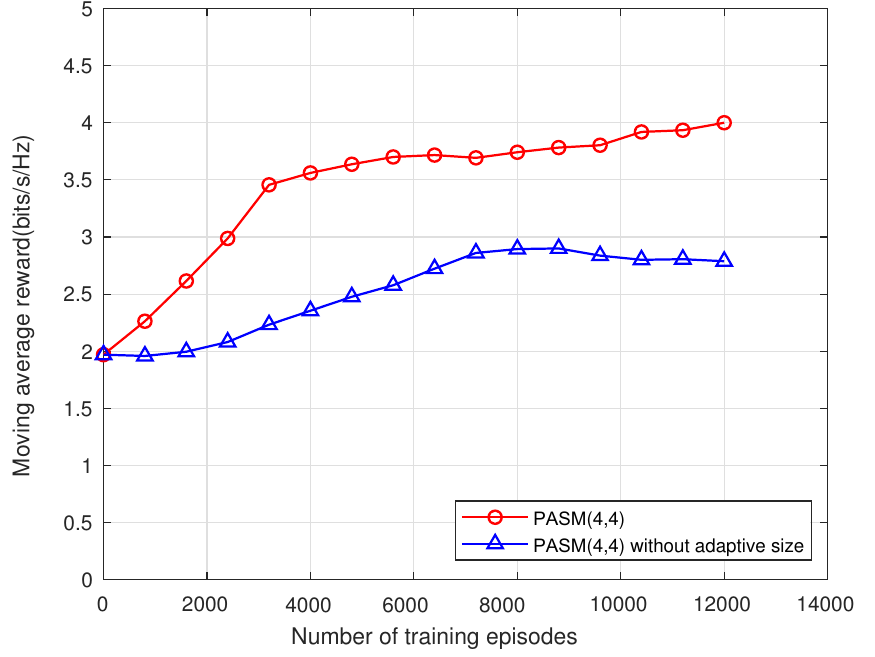}
			\caption{Moving average reward during the training phase. left: Scenario I; Right: Scenario II.}\label{sgd_adapt}
		\end{figure}
		\subsection{Effect of using second moments}
		We first verify the effectiveness of the adaptive stepsize of our algorithm.
		We compare the moving average reward during the training phase of our proposed algorithms with and without adaptive stepsize in both scenarios with $(N,K)=(4,4)$.
		As shown in Fig. \ref{sgd_adapt}, the introduction of adaptive stepsize accelerates the convergence speed in Scenario I while it improves the performance in Scenario II.
		This is similar to the case of the ADAM and RMSprop optimizers, whose effectiveness has been approved in many works in both AI and communication communities.
		Therefore, we only compare our PASM algorithm with adaptive stepsize with other baselines in the following simulations.

		\begin{figure}[th]
			\centering
			\includegraphics[width=0.5\textwidth]{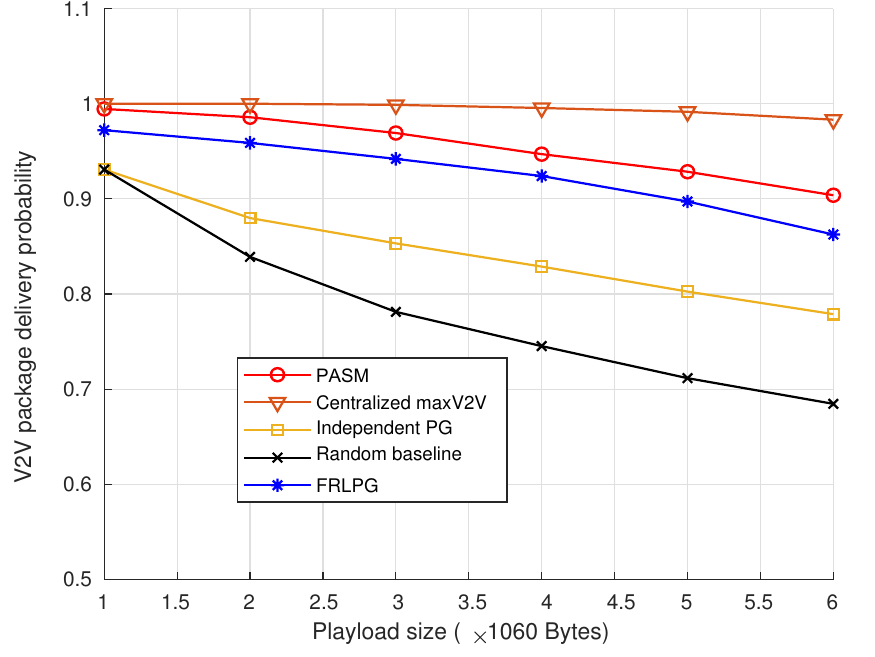}
			\caption{Moving average reward of three algorithms during the training phase.}
			\label{playload}
		\end{figure}
		\subsection{Scenario I}
		Next, we show the experiment results in Scenario I.
		We train the agents for 12000 episodes with $(N,K)=(4,4)$ and a playload size of $2120$ Bytes and then test them in another testing environment.
		To test the robustness of the proposed algorithms to the V2V link playload size, we test the trained model in the environment with $(N,K)=(4,4)$ and increasing V2V playload size.
		Fig. \ref{playload} plots the V2V package delivery rate versus the V2V playload size of the considered algorithms.
		With the playload size increasing, the V2V package delivery rate of all considered algorithms decreases.
		In addition, with any playload size, the agent models trained by our proposed PASM algorithm always have the best testing performance among all algorithms except for the brute-force method, Centralized maxV2V.
		The FRLPG algorithm has better performance than that of the independent PG algorithm.
		When a vehicle moves for a long distance, the environment an agent observes changes significantly.
		With FL manner, the agent is able to learn the new environment from other agents' knowledge, but independent learners cannot.
		Therefore, the Independent PG algorithm has relatively bad performance among all the considered algorithms.
		
		\begin{table}
			\centering
			\caption{Testing performance (V2V link packet delivery rate) of Scenario I with different $(N,K)$ pairs}
			\begin{tabular}{ccccc} 		
				\hline
				(N,K)& PASM &FRLPG&Independent PG&Random baseline\\ 
				\hline 
				$(4,4)$ & \textbf{0.9858} &0.9588 &0.8798 &0.839 \\ 
				$(6,12)$& \textbf{0.9257}& 0.9222& 0.8865 & 0.7797  \\
				$(8,24)$& \textbf{0.8979}& 0.8955&  0.8295& 0.8065\\
				$(6,18)$& \textbf{0.7949}& 0.7439&  0.6893&0.6281\\
				\hline
			\end{tabular}	\label{case1}
		\end{table}
		
		To further test our proposed algorithm, we use the algorithms to train agents in the environments with different $(N,K)$ pairs and then test these trained models in the corresponding environments.
		The $(N,K)$ pair controls the level of training difficulty, as it determines the V2V link density, $N/K$, the number of agents in the environment, $K$, and the freedom degree of resource allocation, i.e., the number of available subchannels, $N$.
		The performance of different algorithms under different $(N,K)$ pairs is summarized in Tab. \ref{case1}.\footnote{We omit the Centralized maxV2V algorithm in this experiment due to its extremely high computational complexity.}
		From the results, the proposed PASM algorithm always has the best performance.
		In addition, as we have explained above, the FRLPG algorithm has better performance than the Independent PG algorithm due to the FL manner among the agents.
		These results validate the efficiency of our proposed PASM algorithm in Scenario I.

		\begin{figure}[th]
			\centering
			\includegraphics[width=0.5\textwidth]{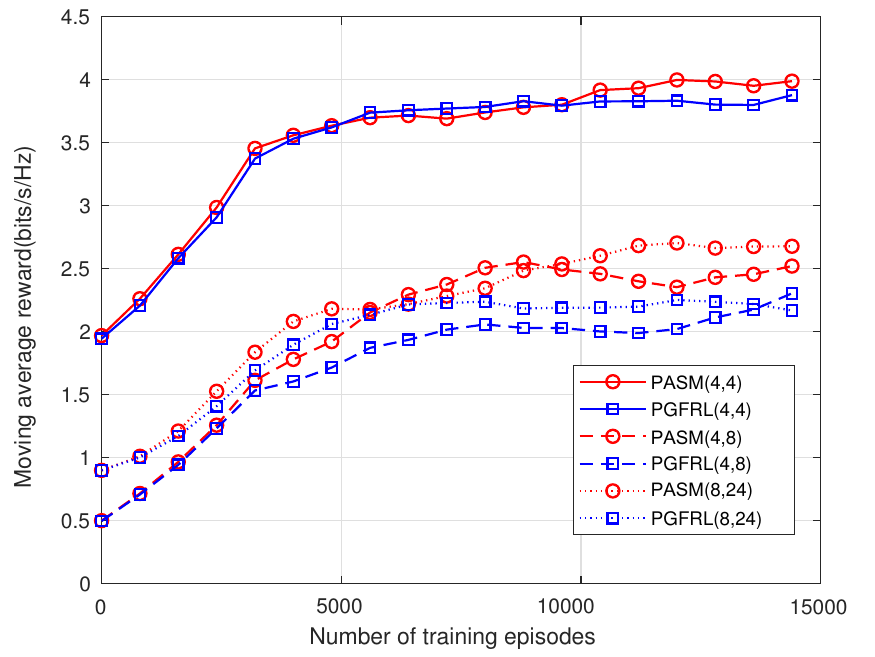}
			\caption{V2I sum-rate and V2V delivery rate  of 3 algorithms.}
			\label{sys_performance}
		\end{figure}
		
		\subsection{Scenario II}
		In the sequel, we evaluate our proposed algorithm in Scenario II.
		Fig. \ref{sys_performance} shows the moving average reward of the PASM algorithm and the FRLPG algorithm versus the number of training episodes in the training phase with different $(N,K)$ pairs.
		These results are obtained by training the agents using the corresponding algorithms for $15000$ episodes and testing in another testing environment.
		From the figure, when $(N,K)=(4,4)$, the PASM algorithm and the FRLPG algorithm have similar performance.
		However, when the number of agents and the V2V link density increase, the PASM algorithm has a significant performance gain over the FRLPG algorithm.
		This is because the weighted sum rate problem is relatively simple compared with the case of $(4,8)$ and $(8,24)$ when $(N,K)=(4,4)$.
		The ADMM-based algorithm has better performance than the Fedavg-based algorithm when the problem is highly non-convex.
		The relative training performance gain of the PASM algorithm over the FRLPG algorithm even reaches around $20\%$ when $(N,K)=(8,24)$.
		
		Then we evaluate the corresponding testing performance of the considered algorithms in the testing environment.
		The testing performance of the considered algorithms under different $(N,K)$ pairs is summarized in Tab. \ref{case2}.
		From the table, the Independent PG algorithm outperforms the FRLPG algorithm when $(N,K)=(4,8)$. 
		This is because agents only optimize the system weighted sum-throughput in the current time slot in Scenario II, while agents have to optimize the V2V link package delivery rate in Scenario I, which is obtained after a sequence of decisions.
		Therefore, it is easier for independent learners to learn a good policy in Scenario II than in Scenario I.
		In addition, the Independent PG algorithm allows each agent to keep its own policy instead of a shared policy, which results in a larger degree of freedom.
		The easier problem setting and the larger degree of freedom together contribute to the performance gain of the Independent PG algorithm over the FRLPG algorithm.
		However, due to the induction of Lagrange multipliers, the PASM algorithm can better tackle the non-convexity of the problem and thus always has the best performance under all environmental conditions.
		These results validate the efficiency of our proposed algorithm in Scenario II.

		\begin{table}
			\centering
			\caption{Testing performance (weighted average rate of all links) of Scenario II with different $(N,K)$ pairs}\setlength{\tabcolsep}{1.5mm}{
				\begin{tabular}{ccccc} 		
					\hline
					(N,K)& PASM &FRLPG&Independent PG&Random baseline\\ 
					\hline
					$(4,4)$ & \textbf{4.0} Mbps &3.71 Mbps & 3.17 Mbps&1.77 Mbps \\ 
					$(4,8)$& \textbf{2.59} Mbps& 2.15 Mbps& 2.41 Mbps & 0.91 Mbps  \\
					$(8,24)$& \textbf{2.72} Mbps& 2.16 Mbps &  2.54 Mbps& 1.27 Mbps\\
					\hline
			\end{tabular}}	\label{case2}
		\end{table}
		
		\section{Conclusion}\label{conclusion}
		We developed a PASM learning algorithm based on the framework of FRL. 
		The algorithm was implemented by inexact ADMM and benefited from two critical techniques: the usage of the PG and the second moment of the Lagrange multipliers. 
		The former enabled the agents to gradually improve their policies while the latter enabled an adaptive learning rate to speed up and to improve the training. 
		We implemented PASM  in a V2X network to train the agents in an FL manner to optimize the V2V package delivery rate and the system weighted sum-throughput.
		The numerical experiment has shown that our proposed algorithm can improve the performance of the resource allocation problem in the considered V2X network.
		
		\begin{appendices}
					\section{Proof of Lemma \ref{convergence-lemma}}
\begin{proof}
Let $\alpha:=r_k/(\rho+r_k)$. It follows from \eqref{lambda_update} and \eqref{original_pg} that
\begin{align*}
\la_k^{j+1}&= (1-\alpha)\g_k^j + \alpha\la_k^{j}\\
&= (1-\alpha)\g_k^j + (1-\alpha)\alpha\g_k^{j-1}+\alpha^2\la_k^{j-1}\\
&= \cdots\\
&= (1-\alpha)\sum_{t=0}^j\alpha^t\g_k^{j-t} +\alpha^{j+1}\la_k^0\\
&= (1-\alpha)\sum_{t=0}^j\alpha^t\g_k^{j-t},
\end{align*}
the last equality is from $\la_k^0=0,$ which results in
\begin{align*}
\|\la_k^{j+1}\|_{\infty} &\leq   (1-\alpha)\sum_{t=0}^j\alpha^t\|g_k^{j-t}\|_{\infty} \\
&\leq  (1-\alpha)\sum_{t=0}^j\alpha^t\|g_k^{j-t}\|_{\infty} \\
& \leq  (1-\alpha)(1-\epsilon)\sum_{t=0}^j\alpha^t \\
&= (1-\alpha^j)(1-\epsilon)^2\leq(1-\epsilon).
\end{align*}
Using the above condition $\mathbf{v}_{c}^{0}=0$, and \eqref{moment_est} we have 
\begin{align*}
			\|\mathbf{v}_{c}^{1}\|_{\infty} &\leq  \| \beta\mathbf{v}_{c}^{0}\|_{\infty}+\frac{1}{K}\sum_{k\in\K}(1-\beta)\|{\boldsymbol{\lambda}}_k^{1}\odot {\boldsymbol{\lambda}}_k^{1}\|_{\infty}\\
 &\leq   ( 1-\beta)(1-\epsilon)^2 < (1-\epsilon)^2,
\end{align*}
which further leads to
\begin{align*}
			\|\mathbf{v}_{c}^{2}\|_{\infty} &\leq  \| \beta\mathbf{v}_{c}^{1}\|_{\infty}+\frac{1}{K}\sum_{k\in\K}(1-\beta)\|{\boldsymbol{\lambda}}_k^{2}\odot {\boldsymbol{\lambda}}_k^{2}\|_{\infty}\\
 &\leq  \beta(1-\epsilon)^2 +   (1-\beta)(1-\epsilon)^2 = (1-\epsilon)^2.
\end{align*}
By deduction, we can show that $\|\mathbf{v}_{c}^{j}\|_{\infty}\leq(1-\epsilon)^2$  for any $j=1,2,3,\cdots$.
\end{proof}
			\section{Proof of Theorem \ref{convergence}}
				For convenience, we define some updating gaps as follows,
				\begin{equation}
					\begin{aligned}
						&\Delta \boldsymbol{\theta}_k^{j+1}=\boldsymbol{\theta}_k^{j+1}-\boldsymbol{\theta}_k^{j},~~ \Delta \boldsymbol{\lambda}_k^{j+1}=\boldsymbol{\lambda}_k^{j+1}-\boldsymbol{\lambda}_k^{j},\\
						&\Delta \boldsymbol{\theta}_c^{j+1}=\boldsymbol{\theta}_c^{j+1}-\boldsymbol{\theta}_c^j,~~ \Delta\th_{kc}^{j+1}=\boldsymbol{\theta}_k^{j+1}-\boldsymbol{\theta}_c^{j},\\
						&\Delta\tilde{\th}_{kc}^{j+1}=\boldsymbol{\theta}_k^{j+1}-\boldsymbol{\theta}_c^{j+1}
					\end{aligned}
				\end{equation}
				For notational convenience, we write
				$$\begin{array}{l}
				\sum:=\sum_{k\in\K}.
				\end{array}$$
				Based on the gradient Lipschitz continuity, we have the following descent inequality for a gradient-Lipschitz-continuous function $f(.)$,
				\begin{equation}
					\begin{aligned}
						f(\mathbf{x})-f(\mathbf{y})\leq \nabla f(\mathbf{w})^\top(\mathbf{x}-\mathbf{y})+\frac{l}{2}\|\mathbf{x}-\mathbf{y}\|^2,
					\end{aligned}
				\end{equation}
				where $\mathbf{w}$ can be $\mathbf{x}$ or $\mathbf{y}$, and $l>0$ is the Lipschitz constant. For ${\bf X}=({\bf x}_1,{\bf x}_2,\cdots,{\bf x}_K)$, ${\bf Y}=({\bf y}_1,{\bf y}_2,\cdots,{\bf y}_K)$, if $f({\bf X})$ is Lipschitz continuous, we have
				\begin{equation}\label{sub_L_cont_desc}
					\begin{aligned}
						&f({\bf X})-f({\bf Y})\\
						\leq &\sum \Big(\nabla_{{\bf w}_k} f({\bf W})^\top(\mathbf{x}_k-\mathbf{y}_k)+\frac{l}{2}\|\mathbf{x}_k-\mathbf{y}_k\|^2\Big),
					\end{aligned}
				\end{equation}
				where ${\bf W}=({\bf w}_1,{\bf w}_2,\cdots,{\bf w}_K)$ can be ${\bf X}$ or ${\bf Y}$.
				Similarly, let $\boldsymbol{\Theta}^i:=(\boldsymbol{\theta}_1^i,\boldsymbol{\theta}_2^i,\ldots,\boldsymbol{\theta}_K^i),i=1,2$, by \eqref{L_count} we have,
				\begin{equation}\label{sub_L_cont}
				\begin{aligned}
				\sum\|\nabla_{\th_k} \phi(\boldsymbol{\Theta}^1)-\nabla_{\th_k} \phi(\boldsymbol{\Theta}^2)\|^2\leq l^2\sum\|\boldsymbol{\theta}_{k}^1-\boldsymbol{\theta}_{k}^2\|^2.
				\end{aligned}
				\end{equation}
				
							\begin{proof}
				We prove the first part of Theorem \ref{convergence} by analyzing the gap between two consecutive updates.
				We rewrite the gap as a sum of three parts as follows,
				\begin{align*} 
					&L(\Z^{j+1}) - L(\Z^j) = e_1^{j+1}+e_2^{j+1}+e_3^{j},\\
					&e_1^{j+1}:= L(\TH^{j+1},\LA^{j},\th_c^j)-L(\TH^{j},\LA^{j},\th_c^j),\\
					&e_2^{j+1}:= L(\Z^{j+1})-L(\TH^{j+1},\LA^{j},\th_c^j),\\
					&e_3^{j~~~}:= L(\TH^{j},\LA^{j},\th_c^j)-L(\Z^{j}).
				\end{align*}			
			1)	We first derive the upper bound of $e_1^{j+1}$, which indicates the updating impact of $\boldsymbol{\theta}_k,\forall k$.
				Based on the gradient Lipschitz continuity of $\phi$, we have,
				\begin{align}
						e_1^{j+1}&=-\phi(\boldsymbol{\Theta}^{j+1})+\phi(\boldsymbol{\Theta}^{j})\nonumber\\
						&+\sum\Big((\la_k^j)^\top\Delta \boldsymbol{\theta}_k^{j+1}+\frac{\rho}{2}(\|\Delta\th_{kc}^{j+1}\|^2-\|\Delta\widetilde{\boldsymbol{\theta}}_{kc}^{j}\|^2)\Big)\nonumber\\
						&  \leq  \sum \Big((\tilde{\g}_k^{j+1}+\boldsymbol{\lambda}_k^{j})^\top \Delta \boldsymbol{\theta}_k^{j+1} +\frac{l}{2}\|\Delta \boldsymbol{\theta}_k^{j+1}\|^2\nonumber\\
			\label{e-1-j+1}			&+\frac{\rho}{2}(\|\Delta\th_{kc}^{j+1}\|^2-\|\Delta\widetilde{\boldsymbol{\theta}}_{kc}^{j}\|^2)\Big),
					\end{align}
				where $\tilde{\g}_k^{j+1}:=-\nabla_{\th_k}\phi(\boldsymbol{\Theta}^{j+1})$ and the inequality is from \eqref{sub_L_cont_desc}.  Then, we have,
	\begin{align*}
					p_k^{j+1}&:=(\tilde{\g}_k^{j+1}+\boldsymbol{\lambda}_k^{j})^\top \Delta \boldsymbol{\theta}_k^{j+1} +\frac{l}{2}\|\Delta \boldsymbol{\theta}_k^{j+1}\|^2\nonumber\\
						&+\frac{\rho}{2}(\|\Delta\th_{kc}^{j+1}\|^2-\|\boldsymbol{\theta}_k^{j}-\th_{k}^{j+1}+\th_k^{j+1}-\boldsymbol{\theta}_c^{j}\|^2)\\
						&=\Big(\tilde{\g}_k^{j+1}+\boldsymbol{\lambda}_k^{j}+\rho\Delta\th_{kc}^{j+1}\Big)^\top \Delta \boldsymbol{\theta}_k^{j+1} +\frac{l-\rho}{2}\|\Delta\boldsymbol{\theta}_k^{j+1}\|^2 \\
						&\overset{\eqref{original_pg}}{=}\Big(\tilde{{\g}}_k^{j+1}-{\g}_k^j-r_k\Delta\th_{kc}^{j+1}\Big)^\top \Delta\boldsymbol{\theta}_k^{j+1} +\frac{l-\rho}{2}\|\Delta\boldsymbol{\theta}_k^{j+1}\|^2.
					\end{align*}					
				Then using two facts $2{\bf a}^\top{\bf b} \leq t\|{\bf a}\|^2+(1/t)\|{\bf b}\|^2$ for any ${\bf a}, {\bf b}$ and $t>0$ and $r_k\in(0,\ell]$, we have 
				\begin{align*}
					p_k^{j+1}&\leq \frac{1}{2l}\|\tilde{{\g}}_k^{j+1}-{\g}_k^j-r_k\Delta\th_{kc}^{j+1}\|^2 
					 +\frac{2l-\rho}{2}\|\Delta\boldsymbol{\theta}_k^{j+1}\|^2\\
					 &\leq \frac{1}{l}\|\tilde{{\g}}_k^{j+1}-{\g}_k^j\|^2+\frac{r_k^2}{l}\|\Delta\th_{kc}^{j+1}\|^2 
					 +\frac{2l-\rho}{2}\|\Delta\boldsymbol{\theta}_k^{j+1}\|^2\\
					 &\leq \frac{1}{l}\|\tilde{{\g}}_k^{j+1}-{\g}_k^j\|^2+ \frac{l^2}{l}\|\Delta\th_{kc}^{j+1}\|^2 
					 +\frac{2l-\rho}{2}\|\Delta\boldsymbol{\theta}_k^{j+1}\|^2\\
					  &= \frac{1}{l}\|\tilde{{\g}}_k^{j+1}-{\g}_k^j\|^2+{l}\|\Delta\th_{kc}^{j+1}\|^2 
					 +\frac{2l-\rho}{2}\|\Delta\boldsymbol{\theta}_k^{j+1}\|^2.
					\end{align*}		
				Therefore, from \eqref{e-1-j+1} and the above condition	we derive 
					\begin{align}
						&e_1^{j+1} \leq \sum p_k^{j+1}\nonumber\\
						\leq &\sum  \Big(\frac{1}{l}\|\tilde{{\g}}_k^{j+1}-{\g}_k^j\|^2+{l}\|\Delta\th_{kc}^{j+1}\|^2 
					 +\frac{2l-\rho}{2}\|\Delta\boldsymbol{\theta}_k^{j+1}\|^2\Big)\nonumber\\				
						\overset{\eqref{sub_L_cont}}{\leq} &\sum  \Big(2{l}\|\Delta\th_{kc}^{j+1}\|^2 
					 +\frac{2l-\rho}{2}\|\Delta\boldsymbol{\theta}_k^{j+1}\|^2\Big)\nonumber\\
						\overset{\eqref{lambda_update}}{=}&\sum  \Big(\frac{2l}{\rho^2}\|\Delta\boldsymbol{\lambda}_k^{j+1}\|^2+\frac{2l-\rho}{2}\|\Delta\boldsymbol{\theta}_k^{j+1}\|^2\Big).
					\end{align}
2) For $e_2^{j+1}$ regarding the impact of updating $\boldsymbol{\lambda}_k,\forall k$, we have
				\begin{equation}
					\begin{aligned}
						e_2^{j+1}=\sum (\Delta\th_{kc}^{j+1})^\top\Delta\boldsymbol{\lambda}_k^{j+1} 
						\overset{\eqref{lambda_update}}{=}\frac{1}{\rho}\sum\|\Delta\boldsymbol{\lambda}_k^{j+1}\|^2.
					\end{aligned}
				\end{equation}								
3) We next derive the upper bound for $e_3^{j}$  about  $\boldsymbol{\theta}_c$. For simplicity, let ${\bf w}_c^{j}:=\sqrt{\mathbf{v}_{c}^{j}}+\epsilon$. By Lemma \ref{convergence-lemma}, we have $ \|{\bf w}_c^{j}\|_{\infty}\in[\epsilon,1)$. 
				Based on the updating steps \eqref{agg_step} and \eqref{new_pg}, we have the following equation,
				\begin{equation*} 
					\sum  \Big(\Delta\widetilde{\boldsymbol{\theta}}_{kc}^{j}+ \frac{1}{\rho{\bf w}_c^{j}}\odot\boldsymbol{\lambda}_k^{j}\Big)=\mathbf{0},
				\end{equation*}
				which immediately results in
					\begin{equation}\label{opt-cond}
					\sum  -\boldsymbol{\lambda}_k^{j} = \sum  \rho{\bf w}_c^{j}\odot\Delta\widetilde{\th}_{kc}^{j}.
				\end{equation}
				It follows from \eqref{lagrange_function} and the above condition that 
				\begin{align}
\label{agg_lambda}						e_3^{j}:=& \sum \Big( -(\la_k^{j})^\top \Delta \th_c^{j}  +\frac{\rho}{2}(\|\Delta\widetilde{\th}_{kc}^{j}\|^2-\|\Delta\th_{kc}^{j}\|^2) \Big) \\
							 =& \sum \Big( \rho({\bf w}_c^{j}\odot \Delta\widetilde{\th}_{kc}^{j})^\top \Delta \th_c^{j} +\frac{\rho}{2}( \|\Delta\widetilde{\th}_{kc}^{j}\|^2-\|\Delta\th_{kc}^{j}\|^2 )\Big).\nonumber
						\end{align}
						The following part aims to estimate the right-hand side of \eqref{agg_lambda}.  
					\begin{align}
						q_k^{j}:=& \rho({\bf w}_c^{j}\odot \Delta\widetilde{\th}_{kc}^{j})^\top \Delta \th_c^{j} + \frac{\rho}{2}\Big( \|\Delta\widetilde{\th}_{kc}^{j}\|^2-\|\Delta\th_{kc}^{j}\|^2 \Big)\nonumber \\
					=&\rho({\bf w}_c^{j}\odot\Delta\widetilde{\th}_{kc}^{j} )^\top \Delta \th_c^{j} -\rho(\Delta\widetilde{\th}_{kc}^{j})^\top \Delta \th_c^{j} -\frac{\rho}{2}\|\Delta\boldsymbol{\theta}_c^{j}\|^2 \nonumber \\
					=&\frac{\rho}{2}\Big\|\sqrt{{\bf w}_c^{j}}\odot(\Delta\widetilde{\th}_{kc}^{j} + \Delta \th_c^{j})\Big\|^2-\frac{\rho}{2}\|\Delta\boldsymbol{\theta}_c^{j}\|^2\nonumber\\
					-&\frac{\rho}{2}\Big\|\sqrt{{\bf w}_c^{j}}\odot\Delta\widetilde{\th}_{kc}^{j}\Big\|^2-\frac{\rho}{2}\Big\|\sqrt{{\bf w}_c^{j}}\odot \Delta \th_c^{j}\Big\|^2 \nonumber \\
					-&\frac{\rho}{2}\Big\|\Delta\widetilde{\th}_{kc}^{j} + \Delta \th_c^{j}\Big\|^2+\frac{\rho}{2}\Big\|\Delta\widetilde{\th}_{kc}^{j}\Big\|^2+\frac{\rho}{2}\Big\| \Delta \th_c^{j}\Big\|^2 \nonumber \\
					\leq&\frac{\rho(\|{\bf w}_c^{j}\|_\infty-1)}{2}\Big\|\Delta\widetilde{\th}_{kc}^{j} + \Delta \th_c^{j}\Big\|^2-\frac{\rho}{2}\|\Delta\boldsymbol{\theta}_c^{j}\|^2\nonumber\\
					+&\frac{\rho(1-\epsilon)}{2}\Big\|\Delta\widetilde{\th}_{kc}^{j}\Big\|^2+\frac{\rho(1-\epsilon)}{2}\Big\| \Delta \th_c^{j}\Big\|^2 \nonumber \\
						\leq&\frac{\rho(1-\epsilon)}{2}\Big\|\Delta\widetilde{\th}_{kc}^{j}\Big\|^2-\frac{\epsilon\rho}{2}\Big\| \Delta \th_c^{j}\Big\|^2  \nonumber \\
						\leq&\frac{\rho }{4}\Big\|\Delta\widetilde{\th}_{kc}^{j}\Big\|^2-\frac{\rho}{4}\Big\| \Delta \th_c^{j}\Big\|^2 .\nonumber 
					\end{align}
		where the last three inequalities used $ \|{\bf w}_c^{j}\|_{\infty}\in[\epsilon,1)$ and $\epsilon\in[ 1/2,1]$.	By \eqref{lambda_update}, we have $\Delta\widetilde{\th}_{kc}^{j}=\frac{1}{\rho}\Delta\boldsymbol{\lambda}_k^{j+1} - \Delta \boldsymbol{\theta}_k^{j+1}$ and hence 
		\begin{align}
						q_k^{j}&\leq \frac{\rho }{4}\Big\|\frac{1}{\rho}\Delta\boldsymbol{\lambda}_k^{j+1} - \Delta \boldsymbol{\theta}_k^{j+1}\Big\|^2 - \frac{\rho}{4}\|\Delta\boldsymbol{\theta}_c^{j}\|^2 \nonumber\\
						&\leq \frac{1}{\rho} \| \Delta\boldsymbol{\lambda}_k^{j+1}  \|^2 + \frac{\rho}{3} \| \Delta \boldsymbol{\theta}_k^{j+1} \|^2- \frac{\rho}{4}\|\Delta\boldsymbol{\theta}_c^{j}\|^2,\nonumber 
					\end{align}	
					where the second inequality is from  $\|{\bf a}+{\bf b} \|^2\leq (1+t)\|{\bf a}\|^2+(1+1/t)\|{\bf b}\|^2$ for any ${\bf a}, {\bf b}$ and $t>0$, which together with 	\eqref{agg_lambda}  derives			
				\begin{equation}
					\begin{aligned}
						e_3^{j} &= \sum q_k^{j} \nonumber\\ 
						&\leq \sum\Big( \frac{1}{\rho} \| \Delta\boldsymbol{\lambda}_k^{j+1}  \|^2 +\frac{\rho}{3} \| \Delta \boldsymbol{\theta}_k^{j+1} \|^2- \frac{\rho}{4}\|\Delta\boldsymbol{\theta}_c^{j}\|^2\Big). 
					\end{aligned}
				\end{equation}
4) It follows from \eqref{original_pg} and \eqref{lambda_update} that  $\boldsymbol{\lambda}_k^{j+1}=r_k(\boldsymbol{\theta}_c^j-\boldsymbol{\theta}_k^{j+1})-{\g}_k^j$.
				Then, we have,
				\begin{equation}
					\begin{aligned}
					\Delta \boldsymbol{\lambda}_k^{j+1}=	r_k(\Delta\boldsymbol{\theta}_c^j-\Delta\boldsymbol{\theta}_k^{j+1})+ {\g}_k^{j-1} -{\g}_k^j,
					\end{aligned}
				\end{equation}
				which by $r_k\in(0,l]$ and \eqref{sub_L_cont} suffices to 
				\begin{align}
					&\sum\|\Delta \boldsymbol{\lambda}_k^{j+1}\|^2 \\
                 \leq &\sum(3l^2\|\Delta\boldsymbol{\theta}_k^{j+1}\|^2+3l^2\|\Delta\boldsymbol{\theta}_c^j\|^2+3\|{\g}_k^j-{\g}_k^{j-1}\|^2\nonumber)\\[1ex]
			\label{delta_lambda}		 \leq & \sum (3l^2\|\Delta\boldsymbol{\theta}_k^{j+1}\|^2+6l^2\|\Delta\boldsymbol{\theta}_c^j\|^2).
				\end{align}	
5) We sum $e_1^{j+1}, e_2^{j+1}$ and $e_3^j$ and use \eqref{delta_lambda} to obtain
			\begin{align}
						&L(\Z^{j+1}) - L(\Z^j) = e_1^{j+1}+ e_2^{j+1}+e_3^j\nonumber\\
						=&\sum \Big(\frac{ 6l- \rho}{6} \|\Delta \boldsymbol{\theta}_k^{j+1}\|^2 -\frac{\rho}{4} \|\Delta\boldsymbol{\theta}_c^{j}\|^2+   (\frac{2l}{\rho^2}+\frac{2}{\rho})\|\Delta \boldsymbol{\lambda}_k^{j+1}\|^2\Big)\nonumber\\
						\leq&\sum \Big(  \frac{6l^3}{\rho^2}+\frac{6l^2}{\rho}+\frac{6l-\rho}{6}\Big) \|\Delta \boldsymbol{\theta}_k^{j+1}\|^2\nonumber\\
					 +&\sum \Big(  \frac{12l^3}{\rho^2}+\frac{12l^2}{\rho} -\frac{\rho}{4}\Big) \|\Delta\boldsymbol{\theta}_c^{j}\|^2\nonumber\\
\label{itter_err}	\leq&\sum   \Big(-\frac{61l }{150} \|\Delta \boldsymbol{\theta}_k^{j+1}\|^2 -\frac{9l}{50}  \|\Delta\boldsymbol{\theta}_c^{j}\|^2\ \Big).
					\end{align}	
					where the last inequality is due to $\rho\geq 10l$. From \eqref{itter_err}, we can conclude that sequence $\{L(\mathbf{Z}^j)\}$ is a non-increasing.\\	
6) Based on the descent inequality \eqref{sub_L_cont_desc}, we have,
				\begin{equation*}
					\begin{aligned}
						&\phi(\TH^{j+1})-\phi(\TH_c^{j})\\
						\leq &\sum \Big(({\g}_k^j)^\top\Delta\th_{kc}^{j+1} + \frac{l}{2}\|\Delta\th_{kc}^{j+1}\|^2\Big)\\
						=&\sum \Big((\boldsymbol{\lambda}_k^{j+1}+r_k\Delta\th_{kc}^{j+1})^\top\Delta\th_{kc}^{j+1} + \frac{l}{2}\|\Delta\th_{kc}^{j+1}\|^2\Big)\\
						\leq&\sum \Big((\boldsymbol{\lambda}_k^{j+1})^\top\Delta\th_{kc}^{j+1} + \frac{3l}{2}\|\Delta\th_{kc}^{j+1}\|^2\Big),
					\end{aligned}
				\end{equation*}
				where the equality is due to \eqref{lambda_update} and \eqref{original_pg}. This results in
				\begin{equation*}
					\begin{aligned}
						&L(\mathbf{Z}^{j+1})\\
						=&-\phi(\TH^{j+1})+\sum\Big( (\boldsymbol{\lambda}_k^{j+1})^\top\Delta\th_{kc}^{j+1} + \frac{\rho}{2} \|\Delta\th_{kc}^{j+1}\|^2\Big)\\
						\geq &-\phi(\TH_c^{j})+\frac{\rho-3l}{2}\sum \|\Delta\th_{kc}^{j+1}\|^2.
					\end{aligned}
				\end{equation*}
				Therefore, we have $L(\mathbf{Z}^{j+1})>-\infty$ for due to $\rho \geq 10l$. This together with the non-increasing property of $\{L(\mathbf{Z}^j)\}$ shows that $\{L(\mathbf{Z}^j)\}$ is convergent. Then taking the limit of the both sides of  \eqref{itter_err} immediately leads to $\lim_{j\to \infty} \|\Delta \boldsymbol{\theta}_k^{j}\|=0$ and $\lim_{j\to \infty} \|\boldsymbol{\theta}_c^j\|=0$, which by \eqref{delta_lambda}	contributes to $\lim_{j\to \infty}\|\Delta \boldsymbol{\lambda}_k^{j+1}\|=0$ and $\lim_{j\to \infty}\|\Delta \th_{kc}^{j+1}\|=0$   by \eqref{lambda_update}.	\\				
7) Based on \eqref{lambda_update} and \eqref{original_pg}, we have,
				\begin{equation}\label{grad_up}
					\begin{aligned}
						\left\|\sum {\g}_k^j\right\|=&\Big\|\sum (\boldsymbol{\lambda}_k^{j+1}+r_k\Delta\th_{kc}^{j+1})\Big\|\\[1ex]
						\leq & \Big\|\sum  \boldsymbol{\lambda}_k^{j+1}\Big\|+\sum  l\Big\|\Delta\th_{kc}^{j+1}\Big\|.
					\end{aligned}		
				\end{equation}
				For the first term in \eqref{grad_up}, by  \eqref{opt-cond}, we can conclude that
				\begin{align*} 
					\Big\|\sum \boldsymbol{\lambda}_k^{j+1} \Big\|&=\Big\|\sum  \rho{\bf w}_c^{j+1}\odot ( \boldsymbol{\theta}_k^{j+1}-\boldsymbol{\theta}_c^{j+1} )\Big\|\\[1ex]
					&=\Big\|\sum  \rho{\bf w}_c^{j+1}\odot ( \Delta \boldsymbol{\theta}_{kc}^{j+1}-\Delta \boldsymbol{\theta}_{c}^{j+1} )\Big\|\\[1ex]
					&\leq\rho \sum \| {\bf w}_c^{j+1}\|_\infty \|( \Delta \boldsymbol{\theta}_{kc}^{j+1}-\Delta \boldsymbol{\theta}_{c}^{j+1} )\|\\[1ex]					
					&\leq\rho\sum  \Big(\| \Delta \boldsymbol{\theta}_{kc}^{j+1}\|+\|\Delta \boldsymbol{\theta}_{c}^{j+1} \|\Big)\\[1ex]
					&\to0.
				\end{align*}
The above condition, $\lim_{j\to \infty}\|\Delta \th_{kc}^{j+1}\|=0$, and \eqref{grad_up}	show  $\lim_{j\to \infty}\|\sum_k{\g}_k^j\|=0$.
			\end{proof}
		\end{appendices}
		
		\bibliographystyle{IEEEtran}
		\bibliography{bib}
		
	\end{document}